# MoveLight: Enhancing Traffic Signal Control through Movement-Centric Deep Reinforcement Learning

Junqi Shao[1], Chenhao Zheng[1], Yuxuan Chen[2†],
Yucheng Huang[1†], Rui Zhang[1†]

[1]School of Civil Engineering, Tsinghua University, Beijing 100084, P.R. China
[2]School of Vehicle and Mobility, Tsinghua University, Beijing 100084, P.R. China

Contributing authors: shaojq23@mails.tsinghua.edu.cn; zhengch22@mails.tsinghua.edu.cn;
yuxuan-c23@mails.tsinghua.edu.cn; huangyc20@mails.tsinghua.edu.cn; zhang-r21@mails.tsinghua.edu.cn;
†These authors contributed equally to this work.

**Abstract**

This paper introduces MoveLight, a novel traffic signal control system that enhances urban traffic management through movement-centric deep reinforcement learning. By leveraging detailed real-time data and advanced machine learning techniques, MoveLight overcomes the limitations of traditional traffic signal control methods. It employs a lane-level control approach using the FRAP algorithm to achieve dynamic and adaptive traffic signal control, optimizing traffic flow, reducing congestion, and improving overall efficiency. Our research demonstrates the scalability and effectiveness of MoveLight across single intersections, arterial roads, and network levels. Experimental results using real-world datasets from Cologne and Hangzhou show significant improvements in metrics such as queue length, delay, and throughput compared to existing methods. This study highlights the transformative potential of deep reinforcement learning in intelligent traffic signal control, setting a new standard for sustainable and efficient urban transportation systems.

## 1 Introduction

Transportation is the "circulatory system" of the socio-economic structure, and humanity has long recognized the significant role transportation plays in the economy. The efficiency of a modern socio-economic system is determined by the speed and quality of its flows of people, goods, energy, information, and capital. Transportation serves as one of the main carriers of these "five flows," forming the foundation for the existence and development of modern economic society [1]. With the continuous advancement of social productivity and human progress, urban populations have rapidly increased, and the number of cars in cities has surged. However, the development of urban infrastructure has been limited, leading to increasingly severe urban traffic congestion. This congestion brings about substantial pollution emissions and energy waste, seriously affecting further urban development. Therefore, alleviating traffic congestion has become a pressing issue. Traffic signal control can help us maximize the use of existing resources without building new roads, achieving the goal of improving traffic efficiency at a lower cost.

The development of traffic signal control has undergone significant evolution from simple fixed-timing systems to advanced intelligent control systems. Initially, fixed-timing control operated at preset intervals, lacking adaptability to real-time traffic conditions. Early classic control algorithms, such as the Webster algorithm, improved intersection efficiency by calculating optimal cycle times. With technological advancements, actuated control systems emerged, utilizing ground sensors to detect vehicles and dynamically adjust signal timings.

To enhance traffic signal efficiency, adaptive control systems such as Split Cycle Offset Optimizing Technique(SCOOT) and Sydney Coordinated Adaptive Traffic System(SCATS) were developed, dynamically adjusting signal timings based on real-time traffic data. Meanwhile, model-based optimization methods like the MaxBand algorithm optimized signal coordination to maximize the green wave on main roads, improving vehicle throughput. The Max-Pressure algorithm, on the other hand, optimized traffic flow distribution and signal timings by allocating pressure values.

Subsequently, model-based optimization methods, using traffic flow models and optimization algorithms such as linear programming and dynamic programming, further enhanced traffic efficiency on a larger scale. In recent years, with the advancement of big data and computational power, machine learning and data-driven approaches have been widely applied in traffic signal control. Particularly, reinforcement learning (RL) technology, through trial-and-error learning, enables agents to find optimal signal control strategies. Applications of RL range from optimizing single intersections to coordinating multiple intersections. The integration of various technologies has led to the creation of Intelligent Transportation Systems (ITS), combining sensor networks, IoT, big data analytics, and artificial intelligence to achieve comprehensive traffic management and optimization.

However, in existing research, most studies simply consider four-phase and eight-phase intersections with very coarse phase designs. They also overlook the rapid development and proliferation of roadside sensing devices, as well as the detailed data these devices can detect and provide. Therefore, to fill a gap in the current research in this field, we propose a lane-level traffic signal control method driven by fine-grained data. By utilizing detailed real-time data from each lane, we can achieve higher precision control by optimally combining lanes with different traffic directions. This approach has been applied and validated at the single intersection, arterial, and network levels, demonstrating the scalability of the model.

The remainder of the paper is organized as follows. Section 2 reviews the research on reinforcement learning in the field of traffic light control. Section 3 explains and defines the research problem. Section 4 describes the research methodology. Section 5 presents and analyzes the experimental results. Finally, conclusions and recommendations are delivered in Section 6.

## 2 Related Work

In recent years, reinforcement learning (RL) has emerged as a powerful approach for traffic signal control (TSC) at both individual intersections and larger urban networks. Traditional methods for TSC, such as fixed-time and actuated control, often fall short in dynamically changing traffic environments due to their limited adaptability and reliance on pre-defined rules. Adaptive signal control methods, although more responsive, still struggle to fully optimize traffic flow under fluctuating conditions. Reinforcement learning offers a promising alternative by enabling systems to learn optimal control policies through interaction with the traffic environment, adapting to real-time changes more effectively than traditional methods.

For single intersections, various RL-based approaches have been proposed and demonstrated significant improvements in traffic management. For example, Prashanth & Bhatnagar et al. [2] employed RL with average cost optimization to adaptively control traffic lights, demonstrating improvements over traditional fixed-time control methods. Gao et al.[3] utilized deep Q-learning with experience replay and target networks to optimize signal timings at single intersections, showing substantial performance gains in simulation environments. Nishi et al.[4] applied RL with graph convolutional neural networks (GCNs) to capture spatial dependencies between lanes, leading to more effective traffic signal policies. Additionally, decentralized control strategies, where each intersection operates independently but benefits from shared learning frameworks, have been explored by researchers like Liang et al. [5], showing promising results in managing traffic more efficiently.

When it comes to multi-intersection or network-level TSC, the complexity increases significantly due to the interactions between adjacent intersections. Multi-agent reinforcement learning (MARL) techniques have been developed to handle such scenarios, where multiple RL agents cooperate or compete to optimize traffic flow across a network. In these systems, each intersection is typically controlled by an individual RL agent, and the agents communicate to coordinate their actions. Studies have shown that MARL can effectively manage traffic in urban networks, reducing congestion and improving travel times across multiple intersections. Research by Wei et al.[7] and Noaeen et al.[8] systematically reviewed the applications of MARL in urban networks, demonstrating the potential for scalable and efficient TSC systems that leverage the cooperation between agents to optimize network-wide traffic performance.

One notable challenge in applying RL to TSC is the high dimensionality and dynamic nature of the traffic environment. Techniques such as deep reinforcement learning (Deep RL) have been employed to address these challenges by using deep neural networks to approximate the value functions or policies, enabling the handling of more complex and high-dimensional state spaces. Recent advances in Deep RL have shown promising results in both single intersection and network-level TSC, as highlighted in the systematic literature review by Noaeen et al.[8], which underscored the significant performance gains achieved through the use of advanced Deep RL methods in urban traffic networks.

Several case studies have validated the practical applicability of RL-based traffic signal control. For instance, Li et al.[9] implemented deep reinforcement learning to optimize signal timing in a busy urban intersection, achieving better performance compared to fixed-time and actuated control systems. Furthermore, Schutera et al.[10] demonstrated the scalability and robustness of distributed deep RL in real-world traffic networks, effectively managing complex traffic patterns. Despite these advancements, several issues remain to be addressed before RL-based TSC can be widely deployed in real-world scenarios. These include computational complexity, the need for extensive training data, and the development of robust models that can generalize well across different traffic conditions. Future research is directed towards overcoming these challenges by developing more efficient algorithms, improving the scalability of RL approaches, and integrating RL with other advanced traffic management techniques.

Overall, the integration of RL in both single intersection and network-level TSC represents a significant advancement in the field of intelligent transportation systems. This study builds on existing reinforcement learning algorithms and proposes lane-level traffic signal control based on refined data. Innovations and designs are applied to the action space and state space, effectively avoiding the issue of high dimensionality. Additionally, the method is adaptively extended to both single intersection and multi-intersection scenarios.

## 3 Problem Definition

**Traffic Movement:** A traffic movement is defined as the process of vehicles traveling from an entering lane through the intersection to an exiting lane. We denote a traffic movement from road $l$ to road $m$ as $(l, m)$. In Figure 1, each direction has three possible traffic movements: straight, left turn, and right turn. The state space size is $4 \times n^4$, where the current signal status is a binary variable, and the lane queue capacity is $n$.

Signal Phase

**Traffic signal phase:** $s$ is defined as a set of permissible traffic movements. Figure 2 shows the configuration of signal phases.

In this example, the intersection has eight phases, each allowing different directions of traffic to proceed. The state space size for signal phases is $2^8 \times n^8$.

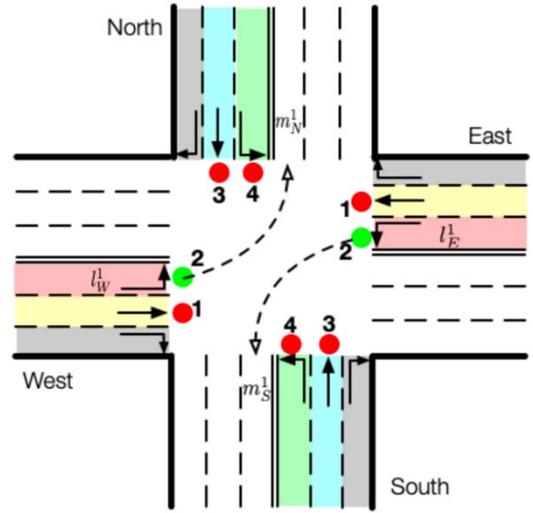

Figure 1 Six-lane intersection movement path

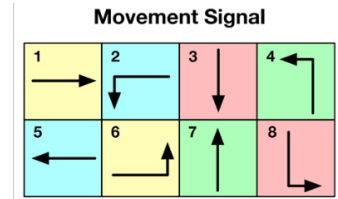

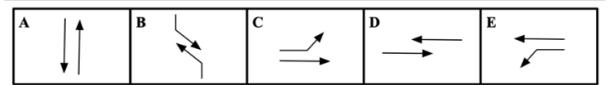

Figure 2 Movement signal

**Pressure of Each Signal Phase:** For each signal phase $s$, there are several permissible traffic movements $(l, m)$. The discrepancy of the number of vehicles on lane $l$ and lane $m$ for traffic movement $(l, m)$ is denoted by $z(l, m)$. The pressure of a signal phase $p(s)$ is simply the total sum of the pressure of its permissible phases $\sum_{(l,m)} z(l, m), \forall (l, m) \in s$.

**Multi-Intersection Traffic Signal Control:** In this section, we define the traffic signal control problem as a Markov game. Each intersection in the system is controlled by an agent. The agents observe a portion of the system state and decide which signal phase to choose at each intersection to minimize the average queue length on the lanes.

Policy and Discount Factor: The actions taken by the agents have long-term effects on the system, so the goal is to minimize the expected delay at each intersection in each episode. At time $t$, each agent selects an action according to the policy $\pi$, aiming to maximize its total reward $G_t^i$. The total reward $G_t^i$ is approximated by neural networks by minimizing the following loss:

$$\mathcal{L}(\theta_n) = \mathbb{E}\left[(r_t^i + \gamma \max_a Q(o_{t+1}^i, a_{t+1}^i; \theta_{n-1}) - Q(o_t^i, a_t^i; \theta_n))^2\right]$$

where $o_{t+1}^i$ denotes the next observation for $o_t^i$. These earlier snapshots of parameters are periodically updated with the most recent network weights and help increase learning stability by decorrelating predicted and target Q-values.

**Green Wave Control:** Figure 3 illustrates the concept of green wave control, emphasizing coordination between several downstream signals. The state-of-the-art Max Pressure (MP) algorithm provides a greedy solution, often resulting in locally optimal solutions.

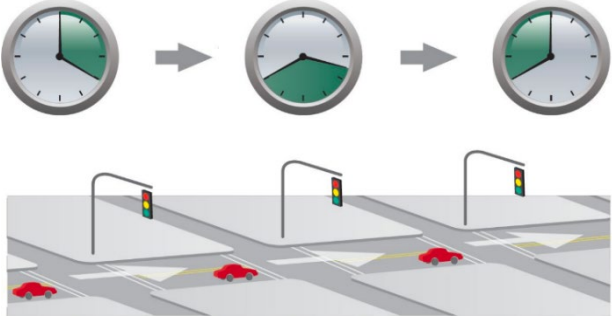

Figure 3 Green wave effect

# 4 Methodology

The research focuses on developing and implementing FRAP (Flip and Rotation and considers All Phase configurations), an advanced algorithm designed for movement-level traffic signal control [11]. This approach leverages deep reinforcement learning (DRL) to create a dynamic and adaptive traffic signal system capable of responding to real-time traffic conditions.

## 4.1 FRAP

FRAP, which stands for Flip and Rotation and considers All Phase configurations, is an innovative algorithm designed to tackle the complexities of movement-level traffic signal control in urban settings [11]. Traditional traffic signal control methods often fall short in adapting to real-time traffic conditions due to their reliance on static models and historical data. FRAP addresses these limitations by leveraging deep reinforcement learning (DRL) to create a dynamic, responsive traffic signal control system that optimizes the flow of traffic based on current conditions.

The FRAP algorithm is built upon the Ape-X DQN (Deep Q-Network) distributed framework. This framework enhances learning efficiency by distributing the computational load across multiple agents. Each agent learns to control traffic signals based on real-time data, improving the system's overall adaptability. Central to FRAP is the phase competition principle, which prioritizes traffic movements with higher demands for green time. This ensures that phases with more vehicles are allocated green lights more frequently, optimizing traffic flow and reducing congestion.

## 4.2 Conflict Matrix for Movements & Phase

FRAP employs a conflict matrix to manage the interactions between different traffic movements. This matrix categorizes movements into conflicting, partially conflicting, and non-conflicting groups, allowing the algorithm to make informed decisions about phase transitions, as shown in Figure 5. Conflicting movements are those that cannot occur simultaneously without causing a collision, such as left turns and opposing through movements. Partially conflicting movements can happen concurrently under certain conditions, like right turns and through movements from the same direction, provided there are no crossing pedestrians. Non-conflicting movements do not interfere with each other and can occur simultaneously without any risk of collision, such as right turns from different directions that do not intersect. By utilizing this conflict matrix, FRAP systematically evaluates which combinations of movements can safely occur during the same phase, optimizing the overall signal timing plan and enhancing traffic flow efficiency while minimizing the risk of accidents [11].

Figure 4 Conflict matrix for movement

## 4.3 Phase Invariant Design

The phase invariant design in FRAP ensures flexibility and adaptability in traffic signal control across various intersection configurations and traffic conditions. This design approach includes several key components. First, phase demand modeling uses neural networks to generate representations of the demand for each traffic phase as shown in Figure 6, incorporating real-time data such as vehicle counts, speeds, and current signal statuses. This accurate modeling of phase demand helps prioritize which phases should be activated. Second, phase pair embedding maps the relationships between different traffic phases using the conflict matrix and demand embeddings, capturing the intensity of demand and the nature of the relationships between phases [11]. This mapping is essential for understanding how different phases interact and compete for green time.

Third, phase pair competition predicts the scores for each phase pair by considering their competitive dynamics. By evaluating which phase pairs have higher demand and lower conflict potential using the embeddings and conflict matrix, the system prioritizes these pairs for green signals, thus optimizing traffic flow.

Lastly, FRAP employs a multi-head attention mechanism to capture the complex interactions between different phases and intersections. This mechanism allows the model to focus on the most relevant aspects of the traffic data, effectively weighing the importance of different movements and phases. By understanding and responding to the dynamic nature of urban traffic, the multi-head attention mechanism significantly enhances the system's decision-making process [11]. The structure of the mechanism is shown in Figure 7.

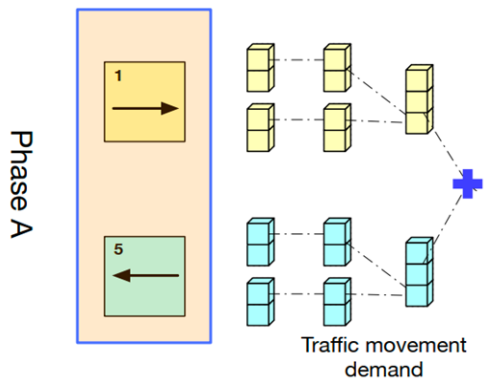

Figure 5 Phase demand modeling

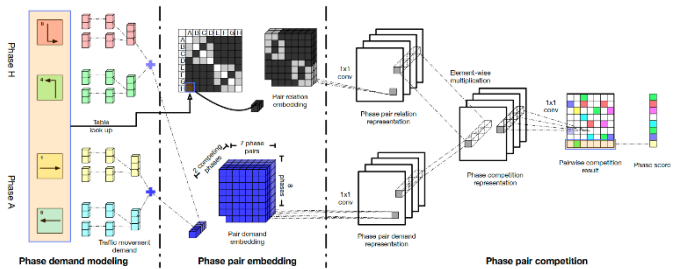

Figure 6 The Phase Invariant Design's structure

### 4.4 Attention and Multi-head Attention

To capture and respond to complex relationships between neighboring intersections, FRAP incorporates a multi-head attention mechanism. This mechanism enables the algorithm to understand and respond to interactions between different intersections, further enhancing the system's ability to optimize traffic flow. By focusing on the most relevant traffic data, the attention mechanism helps FRAP make more accurate and effective control decisions [2].

By assigning higher weights to these critical elements, the model can make more informed decisions about signal timing and phase transitions. The multi-head attention mechanism extends this capability by allowing multiple attention operations to run in parallel, each focusing on different aspects of the traffic situation. For instance, one head might concentrate on the overall traffic flow, while another might focus on congestion at a specific intersection, and yet another might monitor the coordination between adjacent intersections. This parallel processing provides a richer, more nuanced understanding of the traffic environment, enabling FRAP to optimize signal control with greater precision and effectiveness. The result is a robust, adaptive traffic management system that can dynamically respond to real-time traffic conditions, improving overall flow and reducing congestion [2].

### 4.5 Prior Knowledge from Transportation

The FRAP algorithm integrates prior knowledge from transportation science to enhance its traffic signal control capabilities. Inspired by the Max Pressure (MP) algorithm, we denote the pressure of movement $(l, m)$ [7] as:

$$w(l,m) = \frac{x(l)}{x_{max}(l)} - \frac{x(m)}{x_{max}(m)}$$

Our state includes the current phase, the number of vehicles on outgoing lane, and the number of vehicles on each incoming lane. Based on Point Queue theory in traffic engineering, this state definition effectively represents traffic flow with minimal variables. We define the reward as the opposite of intersection pressure, ensuring that our reinforcement learning algorithm can maximize throughput by leveraging the proof established by the Max-Pressure algorithm.

Additionally, the algorithm considers the overall traffic demand and patterns observed from historical data and real-time inputs. This helps FRAP anticipate traffic buildup and dynamically adjust signal timings to prevent congestion before it becomes problematic. The integration of transportation science principles ensures that the system's decisions are grounded in well-established theories and practices, enhancing its reliability and performance.

Moreover, FRAP employs this prior knowledge to create accurate models of traffic behavior, simulating various scenarios to optimize control strategies. This results in traffic signals that maximize flow efficiency, reduce delays, and minimize travel time. By combining machine learning with proven traffic management principles, FRAP offers a robust solution for improving urban mobility and reducing congestion.

# 5 Experiments

## 5.1 Dataset Description

Two real-world datasets, which focus on bi-directional and dynamical flows with turning traffic, are used in our experiments.

**Cologne dataset:** In the experiment, the Cologne dataset is used to evaluate the performance of traffic signal control algorithms. **Road Network**: A 1 × 1 intersection selected from the larger Cologne road network to provide a focused test area for traffic signal control.

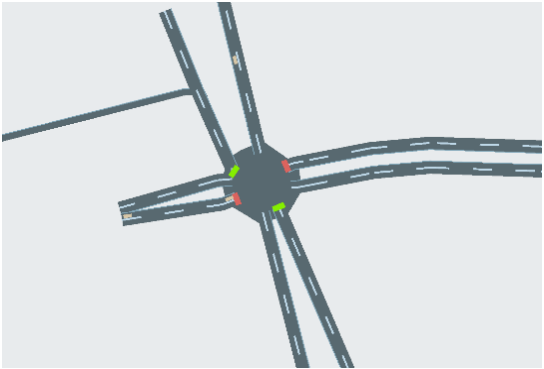

Figure 7 Cologne 1×1 intersection

**Hangzhou dataset:** The Hangzhou dataset is another real-world dataset utilized in the experiments to benchmark traffic signal control algorithms. **Road Network**: Hangzhou 4×4 road network, selected from the Gudang District road network in Hangzhou city.

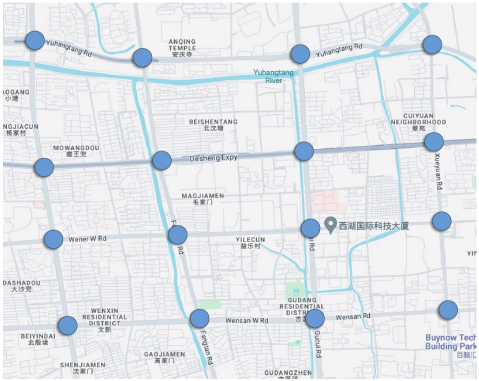

Figure 8 Hangzhou 4×4 real-world network

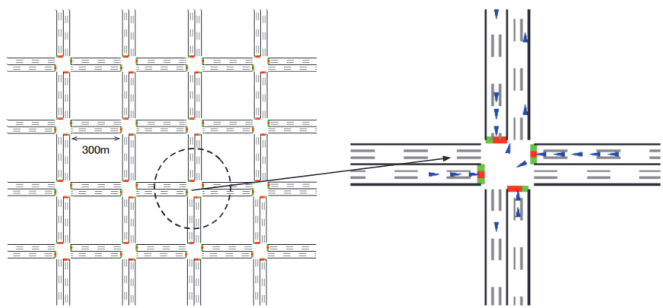

Figure 9 4×4 road network in simulator

## 5.2 Experimental Settings

### 5.2.1 Simulator

Following the previous work on traffic signal control Study, we conduct experiments on Cityflow and SUMO.

**CityFlow** is an open-source traffic simulator that supports large-scale traffic signal control. After the traffic data being fed into the simulator, a vehicle moves towards its destination according to the setting of the environment. The simulator provides the state to the signal control method and executes the traffic signal actions from the control method.

**SUMO** (Simulation of Urban MObility) is an open-source, highly portable, and microscopic traffic simulation software designed to handle large road networks. Developed and maintained by the German Aerospace Center (DLR), SUMO allows for the simulation of traffic dynamics and the evaluation of various traffic management strategies.

For both multi-intersection and single-intersection network setting, we use the real road network to define the network in the simulator.

### 5.2.2 Environmental Settings

**Table 1** Major parameter configuration

| Model parameter | Value |
| --- | --- |
| Learning Rate | 0.001 |
| Batch Size | 64 |
| Buffer Size | 50,000 |
| Training Episodes | 200 |
| Initial Epsilon | 0.8 |
| Epsilon Decay Rate | 0.9995 |
| Minimum Epsilon | 0.01 |
| Test Steps | 3600 |

The model parameters are set as follows: The learning rate is 0.001, which determines the step size during gradient descent. The batch size is set to 64. Buffer size is 50,000, indicating the maximum number of experiences stored for replay. Training consists of 200 episodes to ensure adequate learning. Table 1 details the configuration of these and other major parameters.

Hardware device uses personal computer: Intel(R) Core(TM) i5-13400F 2.50 GHz, Nvidia GeForce RTX 4060 Ti. The version of local virtual machine Ubuntu is 20.04.

Table 2 Performance of all methods

| Metric | Single Intersection | | | | Multiple Intersections | | | |
|---|---|---|---|---|---|---|---|---|
| | Travel time | Queue | Delay | Throughput | Travel time | Queue | Delay | Throughput |
| Webster | 121.8101 | 46.8361 | 5.6219 | 1943 | 534.9851 | 8.8123 | 2.1682 | 2420 |
| MaxPressure | 56.3672 | 10.5139 | 3.3893 | 1996 | 344.9755 | 3.3887 | 0.9964 | 2732 |
| PressLight | 49.8732 | 7.9167 | 0.3948 | 1997 | 329.7855 | 1.6380 | 0.0841 | 2732 |
| CoLight | - | - | - | - | 344.7292 | 1.3554 | 0.0737 | 2733 |
| MPLight | 66.4156 | 17.6833 | 0.5125 | 1994 | 331.1096 | 1.6535 | 0.0907 | 2730 |
| MoveLight | 60.4521 | 11.9917 | 0.4776 | 1997 | 343.0992 | 1.2812 | 0.0735 | 2733 |

### 5.2.3 Evaluation Metric

We select the following four representative measures to evaluate different methods.

**Travel time:** Average travel time of all vehicles in the system is the most frequently used measure to evaluate the performance of the signal control method in transportation.

**Throughput:** It is defined as the number of trips completed by vehicles throughout the simulation. A larger throughput in a given period means a larger number of vehicles have completed their trip during that time and indicates better control strategy.

**Queue Length**: It refers to the number of vehicles lined up at a traffic signal. It is a critical indicator of congestion at intersections. Longer queue lengths typically signify higher congestion levels, leading to increased wait times for vehicles and potential spillback effects.

**Delay**: It measures the additional travel time experienced by vehicles due to traffic signals compared to the travel time in free-flow conditions. Delay is a direct indicator of the inefficiency caused by traffic signals.

### 5.2.4 Compared Methods

We compare our methods with the following baseline methods including both conventional transportation and RL methods. Note that all methods are carefully tuned and their best results are reported.

Conventional transportation baselines:

**Webster**[13]: Webster's Method is a widely used approach in traffic signal control for determining optimal signal timings at intersections. This method focuses on minimizing delays and improving traffic flow efficiency.

**MaxPressure**[14]: the max pressure control selects the phase as green, in order to maximize the pressure according to the upstream and downstream queue length. It is the state-of-the-art control method in the transportation field for signal control in the network level.

RL baselines:

**CoLight**[12]: CoLight represents a cutting-edge approach in traffic signal control, combining the power of deep reinforcement learning with a cooperative multi-agent framework to enhance urban mobility in multiple intersections and reduce traffic-related inefficiencies.

**PressLight**[6]: a recently developed learning-based method that incorporates pressure in the state and reward design fo the RL model. It has shown superior performance in multi-intersection control problems.

**MPLight**[7]: a modern traffic signal control system that utilizes deep reinforcement learning to optimize traffic signal timings at intersections. A distinctive aspect of MPLight is its pressure-based approach

### 5.3 Performance Comparison

### 5.3.1 Performance on Single Intersection

As shown in Table 2, the PressLight method achieved the best performance across various metrics, while our improved method only achieved the best result in terms of throughput.

### 5.3.2 Performance on Multiple Intersection

In this part, we turn to experiments on real-world data. We evaluate our proposed method with other baselines under Hangzhou dataset, where there are $4 \times 4$ intersections. The problem of such a large scale is usually difficult to deal with through conventional methods in the transportation field.

As shown in Table 2, our method achieves the best performance on Queue length, Delay and Throughput. MoveLight showed an improvement in Travel Time of approximately 0.47% compared to CoLight. In terms of Queue, it improved by about 5.47% compared to CoLight and 22.55% compared to MPLight. For Delay, there was no improvement compared to CoLight, but it improved by approximately 18.95% compared to MPLight. The most significant improvements were seen in Queue length and Delay, particularly compared to the Webster and MaxPressure methods. Although there was a slight decrease in Travel Time, MoveLight still performed exceptionally well overall.

### 5.3.3 Impact of Neighbor Quantity and Attention Head Quantity on Performance

**Impact of Neighbor Quantity**

Figure 10 illustrates the effect of neighbor quantity on performance. As the number of neighbors increases from 2 to

5, performance improves and reaches its optimal state. However, further increasing the number of neighbors leads to a decline in performance due to the increased complexity in learning relationships. When determining the signal control strategy for intersections, calculating attention scores for four neighbors and the intersection itself is sufficient to ensure cooperation efficiency and achieve the best performance.

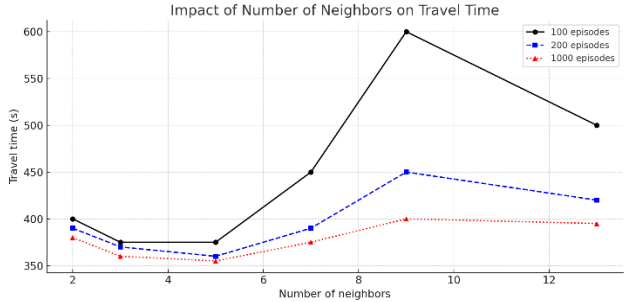

Figure 10 Impact of number of neighbors on travel time across different episodes

**Impact of Attention Head Quantity**

To evaluate the effectiveness of the multi-head attention mechanism, we tested different numbers of attention heads. The results indicate that a moderate number of attention heads can enhance the control of intersection signals. As shown in Table 3, the average travel time of vehicles decreases as the number of attention heads increases. However, the benefit diminishes when the number of heads exceeds 5. Similar conclusions can be drawn from other datasets, though details are omitted due to space constraints.

Table 3 Performance with respect to different numbers of attention heads

| Heads | 1 | 3 | 5 | 7 | 9 |
|---|---|---|---|---|---|
| Travel Time | 177.5 | 174.8 | 172.1 | 175.8 | 175.9 |

The experiments demonstrate that an appropriate number of neighbors and attention heads are crucial factors for improving performance.

### 5.3.4 Running Time

Table 4: Performance of methods in Cologne and Hangzhou dataset

| Metric | Cologne 1×1 | Hangzhou 4×4 |
|---|---|---|
| Webster | 3.3049 | 4.3649 |
| MaxPressure | 2.7601 | 3.1411 |
| PressLight | 3066.7189 | 21630.9526 |
| MPLight | 3148.0649 | 20581.8182 |
| MoveLight | 2641.6128 | 19130.0341 |

Results learning the attention to the neighborhood did not slow down the model's convergence speed; instead, it significantly accelerated the process of reaching the optimal policy. This approach enabled the model to more efficiently assimilate relevant information from its surrounding environment, thereby enhancing its ability to make well-informed decisions. Consequently, the model was able to optimize its performance more rapidly, showcasing the benefits of incorporating attention mechanisms in accelerating the learning and optimization process.

## 6 Conclusion

In this paper, we propose a novel RL method for multi-intersection traffic signal control on the arterials. We conduct extensive experiments using real datasets and demonstrate the superior performance of our method over the state-of-the-art.

We acknowledge the limitations of our model and would like to point out several future directions.

**Environmental Impact Considerations**: As the world faces increasing environmental challenges, incorporating environmental impact assessments into traffic signal control methods will be crucial. Future traffic management systems should be designed with the goal of minimizing vehicular emissions and promoting eco-friendly transportation options. This can be achieved by optimizing signal timings to reduce idling and stop-and-go traffic, which are major contributors to air pollution.

**Incorporation of Multi-modal Transportation**: As urban environments continue to evolve, the future of traffic signal control must extend beyond the traditional focus on vehicular traffic to include a broader range of transportation modes. This includes pedestrians, bicycles, and public transit systems. By integrating multi-modal transportation considerations, traffic management solutions can become more inclusive and equitable.